# RCDN- Robust X-Corner Detection Algorithm based on Advanced CNN Model

Ben Chen, Caihua Xiong*, Quanlin Lim, Zhonghua Wan

*State Key Laboratory of Digital Manufacturing Equipment and Technology, Huazhong University of Science and Technology, Wuhan, Hubei 430074, China*

**Abstract:** Accurate detection and localization of X-corner on both planar and non-planar patterns is a core step in robotics and machine vision. However, previous works could not make a well-balance between accuracy and robustness, which are both crucial criteria to evaluate the detector's performance. To address this problem, in this paper we present a novel detection algorithm which can maintain high subpixel precision on inputs under multiple interference, such as lens distortion, extreme poses and noise. The whole algorithm, adopting a coarse-to-fine strategy, contains a X-corner detection network and three post-processing techniques to distinguish the correct corner candidates, as well as a mixed subpixel refinement technique and an improved region growth strategy to recover the checkerboard pattern partially visible or occluded automatically. Evaluations on real and synthetic images indicate that the presented algorithm has the higher detection rate, subpixel accuracy and robustness than other commonly used methods. Finally, experiments of camera calibration and pose estimation verify it can also get smaller reprojection error in quantitative comparisons to the state-of-the-art.

## 1. Introduction

X-corner is the conjunction point of four adjacent alternating black and white squares, and the corresponding detection is a core step in many machine vision and robotic tasks. For instance, in the field of camera calibration and pose estimation, compared to other fiducial markers (e.g. Deltille [1], ARTag [2], ArUco [3, 4] and circular points [5]), planar chessboard pattern embedded with structured X-corners is widely used due to its distinct, high-contrast features and compact layout [6]. With regard to 3D surface reconstruction [7, 8], projector-camera registration [9] and motion tracking [10], this pattern is also commonly adopted, as X-corner is almost invariant to a certain level of projective transformations and lens distortions [11]. However, the degenerated factors such as sensor noises, large distortion and uneven illumination, can change the appearance of these corners and result in a failed detection. Thus, it is important to design a flexible X-corner detector, which can be flexibly adapted to planar or non-planar pattern and generalized well to most cases.

With respect to the corner detection performance in these tasks, researchers usually focus on two evaluation criteria. The first is the detection accuracy, which consists of detection rate and localization precision. The second criterion is robustness, indicating the detection accuracy can be maintained on low-quality images subject to those degenerated factors above. A reliable detector should perform well in these two aspects so that it can be adopted in wide applications.

There are a mass of detectors attempting to extract valid corner features from various aspects, including the general-purpose detectors[12-14], [15, 16] aided with the well-designed threshold and matching strategies, and the specific detectors [17-20] developed from extracted spatial or frequency properties of X-corner. Some of them may have excellent performance on one or both aspects mentioned above, but rarely strike a good balance among them. Moreover, performance of both general-purpose and specific detectors is easily affected by the extreme poses and apparent deformation. Badly illustration and severe distortion existed in the 3D reconstruction are also the common threats for these detectors' generalization. More details are discussed in the next section.

---

* Corresponding author.
Email address: chxiong@hust.edu.cn

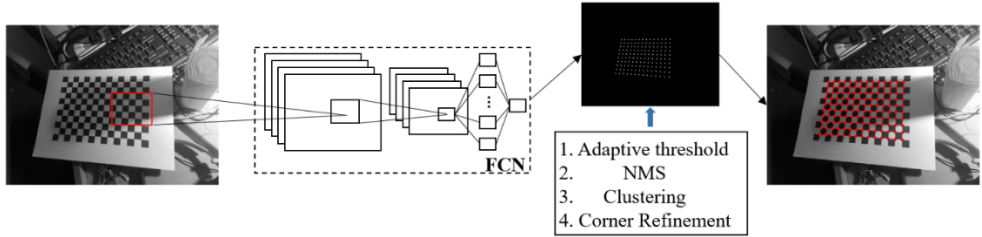

Fig. 1. Overview of the proposed algorithm architecture for accurate X-corner detection using fully convolutional neural networks and some post-processing tips.

CNN has contributed a giant leap in many vision tasks, such as image classification, object detection, semantic segmentation and super resolution reconstruction. The adopted CNN models in these tasks can match or outperform other traditional methods around various criteria without relying on manual features [21]. Naturally we try to use CNN to address the shortcomings mentioned above. But it causes two new challenges. The first is the pixelwise localization precision, unlike in the large-scale object detection tasks that one object usually occupies hundreds or thousands of pixels, the X-true corner location is just one-pixel width. Thus, the corresponding algorithm [22, 23] which generates many anchor boxes and then makes them regress to the ground-truth location, is not suitable for the corner detection. It's because the slight location bias can cause a completely wrong recovery for the pattern with compact X-corners. The other challenge is the extreme foreground-background samples imbalance. For one VGA resolution (640 × 480 pixels) images, the number of true corners is usually less than 100, making the ratio lower than 1:3000, and background points tend to dominate the training result. Thus, it's urgent to propose some new strategies for the detection network's design.

Inspired by the precise pixelwise prediction of the fully convolutional networks (FCN) [24] in semantic segmentation task, here we propose a modified FCN architecture, namely robust X-corner detection network (RCDN), to find the X-corners under multiple cases efficiently. Specifically, we try to address the intensities of each pixel and its surroundings to determine whether it is a corner or not. The outputs in each hidden layer have the same height and width as the input image, which is mainly intended for accurate pixelwise classification. As to the extreme foreground-background samples imbalance, a specific loss function like focal loss [23] is tailored to suppress the vast accumulated loss resulting from the background points. In all, this network can take an image of any size as input and output the response map of same size with a predicted score for each pixel. Subsequently, three post-processing techniques, adaptive threshold, non-maximum suppression and clustering, are used to eliminate false positives for different purposes. We also proposed a new refinement method, which get the subpixel location of the corner by mixing the geometric features from image intensity and final response map. Compared to other techniques, it can get more precise subpixel locations resilient to various bad conditions. Furthermore, derived from the region growth algorithm in [20] and the response map, we present an improved checkerboard recovery strategy. It can automatically recover most checkerboard patterns, whether they are fully visible or partially occluded. In addition, all stages are processed without any prior knowledge or manual interaction.

The outline of this paper is organized as follows. Section 2 introduces the previous related works. Section 3 details the architecture and properties of our checkerboard corners detection network and its training, as well as techniques in the post-processing. Experiments and results are discussed in Section 4. Conclusions are given in the last section.

## 2. Related works

*2.1 X-corner initial detection*

Existing methods based on general-purpose detectors usually employ line fitting to filter out false positives from all candidate points. The typical examples are Hough transform [25, 26] and rotating orthogonal masks convolution [27]. They firstly used Harris detector [12] to find all corners and then distinguished X-corners as the intersections of two groups of grid lines. However, these techniques only work well for low distortion image where the lines should keep straight. Sign-change properties of the gradient circular boundary centered on X-corners [18, 28] can solve this problem, while it can only get good results provided that the environment is not too cluttered. An improved Harris corner detector based on color constancy [29] was proposed to get reliable corner candidates under complicated illumination, and hyperbolic tangent model [30] was utilized to obtain the subpixel corner locations and remove the false positives. But they are still weak in the high distortion or extreme poses.

As for the specific detectors, the widely used method [31] adopted in OpenCV tried to identify black quads and combine them into a complete checkerboard with intersections of their edges extracted as X-corners. It performs well for the images with thick white border and less dark background. An extension of this method implemented in the OCamcalib Toolbox [32, 33] was mainly intended to detect the images with high lens distortion. ChESS [17] is one simple and robust detector with simple subtraction and summation procedures on the sampling circle around each pixel. It is much faster but introduces a lot of background points, and the feasible threshold varies with the image quality. The successor [34] applied machine learning followed ChESS to eliminate false points and gets some improvements. Some works also regarded X-corner as the saddle point, such as Hessian corner detector [35, 36]. They compute efficiently but suffer in noise and perspective transformation. ROCHADE [6] combined complex image processing procedures to build a binary edge graph and extract the saddle points as the X-corners. This method is robust to low resolution and extreme pose, while high distortion and sensitivity to partial shading of light limits its generalization.

Some other specific detectors utilized temple matching to find the target points efficiently. Yu et al. [37] discovered X-corners by 5 double-triangle patterns on the corner patch of the images and computed their correlation coefficient. Geiger et al. [20] matched two types of corner prototypes on each pixel to find the corner likelihood, and expanded the seed points with certain region growth algorithm. OCPAD, an extension of ROCHADE, designed a subgraph matching scheme for partial checkerboard [38]. These detectors are clearly at a disadvantage on image rotation and high distortion.

A preliminary attempt using neural networks is MATE [19], which consists of three convolutional layers to extract the intrinsic feature, but it may cause so many false positives for medium and high-resolution images. Possible reasons are the insufficient model's learning and the lack of postprocessing.

*2.2 Corner refinement methods*

For X-corners there are two typical refinement series: the first series are based on the pixel intensity and can also be divided into two parts: edge approximation techniques used in OpenCV and surface fitting methods [6, 39] used in ROCHADE. The edge approximation techniques using the gradient filters are computed efficiently, but they may lead to severe bias with the noise increasing. The surface fitting methods recognize subpixel corner location as the saddle point of polynomial surface fitted to the surroundings of the corner candidates, and an extension version [40] adopt a Hessian matrix to extract the saddle points for computational efficiency. They can get higher accuracy in most cases but shows less stability against noise and image rotation. The second series are subpixel peak algorithms based on the response map, a detail comparison between different algorithms are given in [41] and the subsection 4.3. They are much faster but the accuracy highly relies on the initial corner detectors.

## 3. Proposed method

In Sec. 3.1 we discuss the overall X-corner detection network architecture, and then detail the specific configurations used in the experiment. Properties of these networks and their training are introduced in Sec. 3.2. Corresponding post-processing techniques and subpixel refinement method are described in Sec. 3.3. The last improved checkerboard recovery strategy will be presented in Sec. 3.4.

### 3.1 Network architecture

A generic layout of the corner detection networks is presented. The basic architecture is the fully convolutional neural network (FCN). The fixed stride of one pixel, and zero padding are adopted in all convolutional layers to make the output equal in size to the input. The activation functions following each convolutional layer are ReLUs. Consider that max-pooling layers result in loss of accurate spatial information [21], we discard them after the convolution.

Kernels of the first layer should be large enough to extract adequate features from the input, as large receptive fields can suppress the effect of focal blur and noise efficiently [17, 19]. MATE found it should be well designed because a larger kernel size may result in lower recall, but a smaller one would cause so many false positives. Consider that the proposed adaptive threshold described later can make a well tradeoff between them. Here we discuss other factors.

First, as a common variant in deep neural networks [22, 23], a stack of smaller filters instead of a single larger one can increase the non-linear rectification and decrease the number of parameters, here we evaluate the effect of this design on the first layer and hidden layers. Second, the size of the last convolutional filters is set to $1 \times 1$. Compared to the former $3 \times 3$, it reduces the computation of non-linear transformations of the response maps across channels without affecting the receptive fields. For some evaluation nets, an extra $1 \times 1$ liner conv. layer was added to fuse the decisions among the previous pretrained models for a better performance. Third, deeper layer has been verified to be useful for large-scale image detection, here we also make a thorough evaluation of networks with increasing depth for searching a best detector.

Based on these discussions, the network configurations evaluated in this paper are outlined in Table 1. Compared to MATE, these nets are deeper, which can extract more features with no significant increase in time consumption, as well as less risk of overfitting.

**Table 1. Network Configuration**[a]

| A | B | C | D | E | F | G | H |
|---|---|---|---|---|---|---|---|
| 3 weight layers | 4 weight layers | 5 weight layers | 4 weight layers | 5 weight layers | 6 weight layers | 6 weight layers | 7 weight layers |
| input (arbitrary size gray image) | | | | | | | |
| conv13-16 | conv13-16 | conv7-16 conv7-16 | conv13-16 | conv13-32 | conv13-32 | conv13-32 | conv13-32 |
| conv1-8 | conv1-8 | conv1-8 | conv3-32 conv3-32 | conv3-32 conv3-32 conv3-32 | conv3-32 conv3-32 conv3-32 | conv5-32 conv3-32 conv3-32 | conv3-32 conv3-32 conv3-32 conv3-32 |
| conv3-1 | conv3-16 conv1-1 | conv3-16 conv1-1 | conv1-1 | conv1-1 | conv1-32 conv1-1 | conv1-32 conv1-1 | conv1-32 conv1-1 |

[a]The convolutional layer parameters are denoted as "conv (receptive field size)-(number of channels)". The ReLU activation function followed each layer is not shown.

### 3.2 Training details

**Dataset.** The datasets for training and evaluation should be large enough with considering various degenerate conditions. In this paper all experiments are performed on two image series: images captured by us directly and digitally augmented versions of them.

For generating abundant training images, we used four checkerboards with different size and different number of X-corners as patterns. Each pattern was placed under various circumstances to capture the datasets, with the background cluttered and illumination blocked intentionally. In order to make our model become rotation and intensity invariant, here we rotated the original images with certain increments and reversed the intensities of half of those pictures randomly. The camera we used has little lens distortion and the capture conditions are good without much noise, so we artificially added both radial and tangential distortion as well as Gaussian noise and blur as mentioned in MATE to multiply these pictures. Finally, all images (a total of 8900) were converted to gray-scale images and resized into VGA resolution with $640 \times 480$ pixels (an optional operation). Among them 8000 images were selected randomly as the training dataset, and the rest were taken as the validation dataset.

X-corner detection is a supervised learning task and the ground-truth pixelwise locations should be annotated accurately. Firstly, we used OpenCV detectors to find all possible corners. For those falsely detected pictures, we indicated these corners with Harris, and then checked and removed all wrong points manually. After annotations the label matrix are created, where $G(x,y)=1$ denotes the real corner and $=0$ for background pixel, thus to some extent the response value can also be regarded as the probability to be a X-corner or not.

**Loss Function.** The extreme imbalance of the number of corners and non-corners is the main obstacle for the network performance. For example, one image with the resolution of 640*480 pixels contains only 56 X-corners, the imbalanced ratio is about 1:5500, much higher than the ratio of 1:1000 for the common large object detection. Thus, the mean loss to each pixel may make the network mistake all of them for the non-corner. Furthermore, for the pixelwise X-corner detection, one more or one less corner falsely detected can even cause large errors to checkerboard's recovery. MATE use the one-side quadratic difference (MSE) with inverse class frequency as the weighting factors to penalize the pixels being wrongly classification. This setting can be feasible for the high-quality images but performance poorly under complicated environment.

Inspired by the success of the focal loss for dense object detection [23], which improves the detection performance by reducing the relative loss for well-classified examples and mainly focusing on the hard examples, here we make adopt it for the pixelwise corner detectors. Collect all trainable parameters into a vector $w$, and take all real corner locations as hard classified examples and the back-ground pixels as easy ones. The total loss function can be defined as:

$$L_{total} = \frac{1}{2}\lambda \| w \|_2^2 - \sum_{(x,y)\in\Omega} \begin{cases} \frac{1}{N_P}(1-a(x,y))\log(a(x,y)), & \text{where } G(x,y)=1, \\ \frac{1}{N_N}a(x,y)\log(1-a(x,y)), & \text{where } G(x,y)=0. \end{cases} \quad (1)$$

where $a(x,y)$ denotes the clipped output of the last layer as:

$$a(x,y) = \begin{cases} \min(\max(10^{-6}, L_6(X)(x,y)), 1), & \text{where } G(x,y)=1, \\ \min(\max(0, L_6(X)(x,y)), 1-10^{-6}), & \text{where } G(x,y)=0. \end{cases} \quad (2)$$

and then the loss function can be meaningful to all pixels' responses. Here inverse class frequency with the number of ground-truth positives ($N_P$) and negatives ($N_N$) is also adopted to reduce the samples disparity. In addition, we use the $L^2$ parameter regularization to reduce the net's overfitting. $\lambda$ is a balancing parameter that weights the contribution of regularization term relative to the focal loss, and here we set $\lambda=0.01$.

**Training.** This network can be trained by back-propagation and stochastic gradient descent (SGD). We use a batch size of 20 images and a momentum of 0.9. The learning rate is initially set to 0.01, and then decreases exponentially with the decay rate of 0.01 as the training progresses. We initialize the weights by using the xaiver initialization method [42], and the neuron biases with the constant 0.1. Our implementation uses TensorFlow.

### 3.3 Techniques for accurate corner localization.

Deciding which responses to be treated as true positives is also worth considering by exploiting spatial and geometric constraints. Here we introduce three efficient techniques designed to eliminate false positive points in different cases, as well as an improved corner refinement method for more accuracy subpixel location.

**Adaptive Threshold.** As the loss function explained above, during the training process our model accelerates responses of corner locations higher than 1 and those of non-corner locations lower than 0. Intuitively, setting a constant threshold of 0.5 is expected to get one much discriminative classification result. However, considered the complex illumination conditions and possible regional blur, which tends to lead the responses subject to different distributions, the global fixed threshold 0.5 (used in MATE) failed to detect all corners in most cases. While regional threshold [17] selects many candidates for each small region, result in the explosion of numbers of false positives. So here we try to address this problem from other views.

In previous work [43] we have ever set the threshold linear with the maximum of responses, based on the assumption that proposed network can make the responses be more discriminative between corners and other pixels at any time, which is proved to be useful for most cases. But the linear factor varies with the input and is hard to decide. Another attempt to separate corners from non-corners is to combine the classification standard with the outputs' quality:

$$T = \mu + \lambda\sigma \qquad (3)$$

where the mean $\mu$ and standard deviation $\sigma$ have been computed for each output and the multiplier factor $\lambda$ gives the level of threshold $T$ depends on the standard deviation. It is called Std-Threshold and performs well for pictures with little noise and skew pose. But this setting is less robust to lens distortions and also introduced an hyper-parameter.

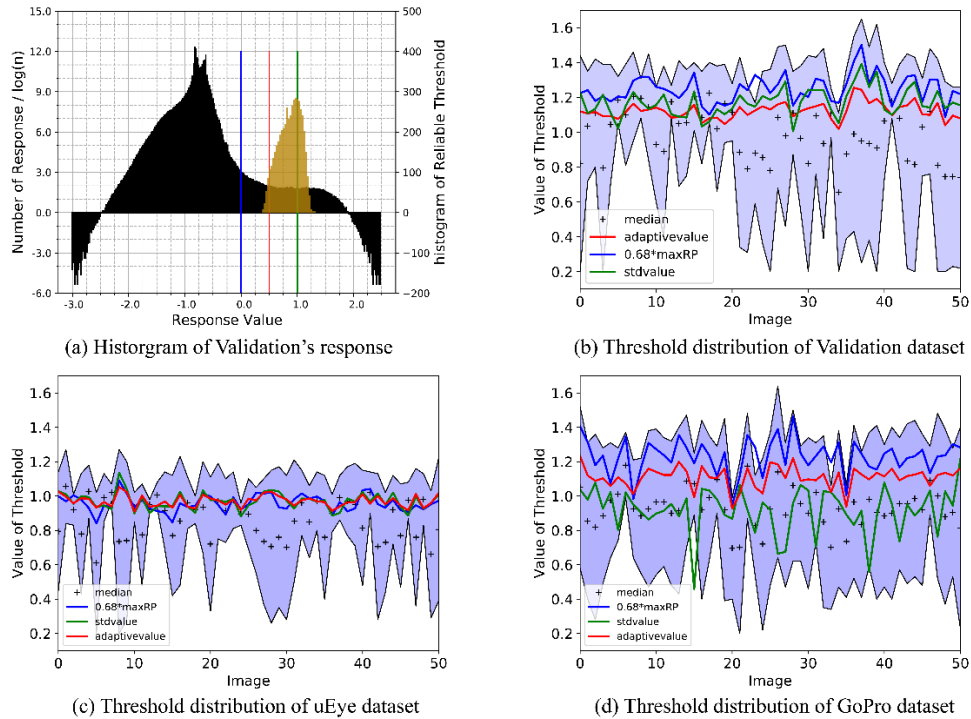

(a) Historgram of Validation's response  
(b) Threshold distribution of Validation dataset  
(c) Threshold distribution of uEye dataset  
(d) Threshold distribution of GoPro dataset  

Fig. 2. Histogram of validations response and adaptive threshold distribution on three datasets.

By observing the output distribution of whole validation images, responses of true corners are nearly all the top $n_{th}$ values (n is the number of X-corners), although several false positives may response high value due to the complex scene. We also find the number of points closer to correct locations and the remaining are approximately equal for those response larger than 0.5 and evenly distributed. Furthermore, the reliable threshold counted on all training and validation images mainly around 0.5-1.5 as shown in Fig. 2 (a). So here we set the threshold as mean of the response values which larger than 0.5. The performance on three datasets are demonstrated in Fig. 2 (b, c and d). We can see this setting can generalizes well to most cases. Also, we will show that the proposed method can get a trade-off between recall and precision.

**Non-maximum Suppression**. Many locations in the immediate neighborhood of corners or near the borders of the image may have slightly lower responses than those corners, and the simpler thresholding fails to eliminate them. So here we use the Non-maximum Suppression (NMS). Firstly, construct bounding-boxes with same size centered around the remaining locations, then apply NMS on them based on the sorted response values, the satisfactory results can be got with the threshold at 0.5. This method is different from those used in previous corner detection algorithm [17, 19] but with high computation efficiency.

**Clustering.** In some complex scenarios, there are many false positives that have very similar appearance to the corners, and their response values are almost the same as those of corners, so that the techniques mentioned above can't distinguish them very well. Considering that the checkerboard has a very regular geometric property, while the false positives are usually distributed randomly and a little away from the checkerboard in the image, we can use the clustering algorithms to separate them. For example in testing we apply k-means++ to the remaining responses with $k=10$, then calculate the number of points $N_i (i=1...$ in each cluster and eliminate points in the cluster with $N_i <= 2$. This tip is valid when testing but can be omitted when the candidates are few (such as less than 30).

**Sub-pixel refinement.** After post-processes above, the remaining can be served as initial corners for the subsequent refinement step. Here we propose a mixed technique which average the refining results of surface fitting method [6] conducted on the pixel intensity and the Gaussian approximation algorithm conducted on the response map. Gaussian approximation algorithm shows excellent efficacy and suitability among the common subpixel peak algorithms [41]. It can be used here to decrease the bias effects of surface fitting method on noise and image rotation with mere marginal cost.

As a side note, the Gaussian approximation algorithm adopts the three highest, contiguous response values around the local-maximum point of the response map and assume the response peak shape fits the Gaussian profile. The subpixel offset of the corner is calculated as:

$$\begin{cases} \delta_x = \frac{1}{2} \frac{\ln(f(x-1,y)) - \ln(f(x+1,y))}{\ln(f(x-1,y)) - 2\ln(f(x,y)) + \ln(f(x+1,y))}, \\ \delta_y = \frac{1}{2} \frac{\ln(f(x,y-1)) - \ln(f(x,y+1))}{\ln(f(x,y-1)) - 2\ln(f(x,y)) + \ln(f(x,y+1))}. \end{cases} \quad (4)$$

and then the refined corner location is $(x+\delta_x, y+\delta_y)$.

### 3.4 Improved checkerboard recovery strategy

For calibration and pose estimation task, the full checkerboard pattern is often used for its simplicity, while for some complicated applications such as face geometry reconstruction and pulmonary function testing [44], the object surface is often not planar and even not fully visible, making the pattern recovery more challenging. A widely used recovery method is the region-growth strategy [21], it can find multiple checkerboards in one single image and is robust to various degenerated factors. But it can only find the fully visible boards and time-consuming.

Firstly, we sort all corners by their response values, and count the edge orientations of the corners with highest value as the local maximums of the gradient histogram, which is computing on the $20 \times 20$ square neighborhood centered on the corner. If the number of these orientations is between 2 and 4, we would calculate adjacent corners along these orientations, otherwise we will discard the initial corners and repeat this process for the corner with second largest value. Then we set each adjacent corner as the initial corner to find their adjacent corners. After that we can get a 2 × 2 or 3 × 3 initial corner indexing matrix, as shown in Fig. 3(a). When the initial matrix width is 3, we extend it by following the region growth method in [21]. But if the width is 2, we will take the boundary corners of the matrix to construct other initial indexing matrixes, and only when the average length of each matrix is between 0.8-1.2 times of the average length of the initial matrix do we combine them. Each step of the improved pattern recovery strategy is illustrated in Fig. 3(b). Those modifications can find the checkerboards partially occluded or not fully visible. Fig. 3(c) shows some excellent detection results on non-planar patterns, which indicates our detector is competent for these wider applications.

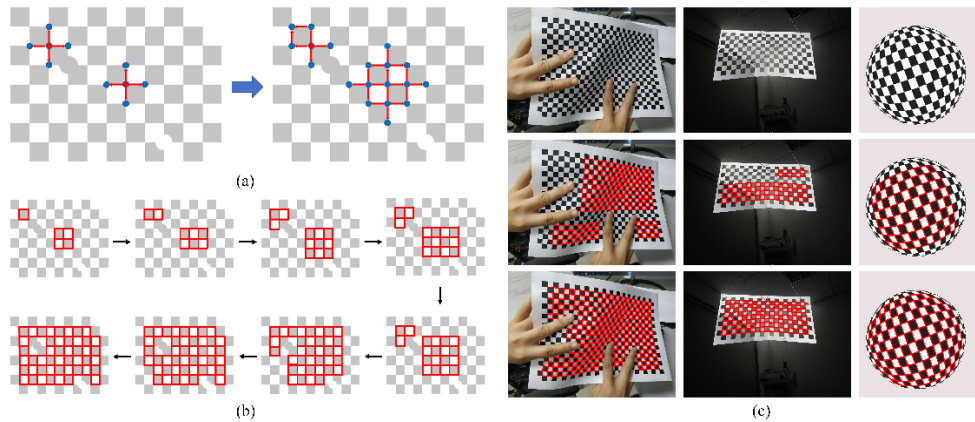

Fig. 3. (a) initial corner indexing matrix; (b) individual steps of improved pattern recovery strategy; (c) detection results on non-planar X-patterns surfaces. The top row shows original images, the second row shows results of region growth [21] and the last of improved methods.

## 4. Experiments

The full evaluations of the proposed X-corner detector are presented in this section. In Sec 4.1, we focus on the performance of network configurations on realistic images to find the best model. For investigate the detector's robustness without any refinements, we apply it as well as several existing algorithms on synthetic image and compare their pixelwise localization error on Sec 4.2. The subpixel refinement accuracy is evaluated in Sec 4.3. Sec 4.4 demonstrate the detector's excellent performance on camera calibration and pose estimation. Finally, we discuss the extending application of proposed detector on other types of corners in the last subsection.

*4.1 Performance on realistic images*

Here we investigate the effects of network configurations on detector's precision and recall in three online available datasets [6]: the Mesa dataset with low-resolution and large amount of noise, uEye dataset with medium resolution and insignificant distortion, and GoPro dataset with high-resolution and strong lens distortions. For a fair comparison, parameters of each model are optimized to result in the overall best performance and keep fixed throughout the evaluation.

In the first experiment, we verify the performance of adaptive threshold and compare the learning ability of proposed loss function with mean squared error (MSE) mentioned in MATE. The bias between the detected corner and the closest ground-truth of all images is calculated, and if the bias is less than four pixels, the detected point is regarded as a true corner.

As shown in Table 2. The fixed threshold 0.5 is a common setting but it tends to sacrifice precision for higher recall due to the complex scenes. and this disparity become worse to the Std-threshold, for multiplier factor $\lambda$ varies largely with input. Threshold related to maximum can get the best results for two datasets, but the linear factor is also a hyper-parameter hard to decide. Compared to the solid threshold of 0.5, we can see that the adaptive threshold can make a well balance between them, and the effect is close to Threshold related to maximum. Its excellent performance can be further shown in later experiments. Furthermore, our loss function performances better than MSE for nearly all threshold methods, indicating it is a more discriminating design. So that in further evaluations we only adopt this improved loss function.

Table 2. Average detection rates at different threshold methods[a]

| Network configuration | Threshold method | Mesa SR4000 | | IDS uEye | | GOPRO Hero3 | |
|---|---|---|---|---|---|---|---|
| | | precision | recall | precision | recall | precision | recall |
| A+ Entropy Cross | 0.5 | 0.9675 | 1.0000 | 0.9296 | 0.9996 | 0.9452 | 0.9996 |
| | Std-(0.8) | 0.9996 | 0.9980 | 0.9987 | 0.9942 | 0.5123 | 1.0000 |
| | 0.41*maximum | 0.9992 | 0.9988 | 0.9961 | 0.9976 | 1.0000 | 0.9983 |
| | Adaptive | 0.9996 | 0.9977 | 0.9993 | 0.9899 | 1.0000 | 0.9970 |
| A+ MSE | 0.5 | 0.9831 | 1.0000 | 0.9263 | 0.9995 | 0.9881 | 0.9996 |
| | Std-(1.7) | 0.9992 | 0.9960 | 0.9939 | 0.9937 | 0.1703 | 1.0000 |
| | 0.57*maximum | 0.9973 | 0.9985 | 0.9959 | 0.9957 | 1.0000 | 0.9963 |
| | Adaptive | 0.9988 | 0.9970 | 0.9982 | 0.9850 | 1.0000 | 0.9959 |

[a]The Std-threshold is denoted as "Std-(multiplier factor value)", and the threshold related to maximum is denoted as "linear factor*maximum". "Adaptive" denotes the proposed adaptive threshold. Them are set for brevity.

The second experiment is designed to evaluate the performances of different network configurations. Here we additionally set model D-3 (the last kernel size is $3 \times 3$ ), E-16 (the first kernel channel is 16) for the quantitative comparison. The detect results of MATE and ChESS are also shown as they are the perfect X-corner detectors without the prior knowledge. Note that the threshold of the testing ChESS detector varies with the datasets (Mesa SR4000: 0.0, IDS uEye: 0.6, GOPRO Hero 3: 0.45) for the best performance. Results are shown in Table 3.

Table 3. Average detection rates for different network configurations

| Network configuration | Mesa SR4000 | | IDS uEye | | GOPRO Hero 3 | |
|---|---|---|---|---|---|---|
| | precision | recall | precision | recall | precision | recall |
| B | 0.9996 | 0.9988 | 0.9979 | 0.9958 | 1.0000 | 0.9980 |
| C | 0.9984 | 0.9975 | 0.9776 | 0.9987 | 1.0000 | 0.9980 |
| D-1 | 0.9996 | 0.9991 | 0.9989 | 0.9961 | 1.0000 | 0.9987 |
| D-3 | 0.9995 | 0.9987 | 0.9984 | 0.9957 | 1.0000 | 0.9983 |
| E-16 | 0.9995 | 0.9987 | 0.9924 | 0.9977 | 1.0000 | 0.9994 |
| E-32 | 0.9984 | 0.9993 | 0.9955 | 0.9991 | 1.0000 | 0.9994 |
| F | **0.9997** | **0.9992** | **0.9986** | **0.9996** | **0.9994** | **0.9996** |
| G | 0.9990 | 0.9993 | 0.9933 | 0.9993 | 1.0000 | 0.9994 |
| H | 0.9996 | 0.9995 | 0.9961 | 0.9987 | 1.0000 | 0.9991 |
| MATE | 0.9988 | 0.9970 | 0.9982 | 0.9850 | 1.0000 | 0.9959 |
| ChESS | 0.9640 | 0.8191 | 0.9884 | 0.9622 | 0.9602 | 0.8267 |

First, we note that using a stack of smaller filters to replace a single large filter in the first layer (B and C) results to the increase of one of both false positives and missing corners, especial for the small Mesa images. It indicates the smaller local receptive fields can't capture enough spatial contexts of corner features. While for the hidden layer (G and H), smaller filter size can perform better than the larger on precision with merely loss on recall. It complies with the idea that more smaller filters with more non-linear rectification can get a better decision.

Second, the results that model D-1 is better than D-3, shows that $1 \times 1$ convolutional filters for compute reduction and information fusion across channels can get a more correct decision than $3 \times 3$ filters. Besides, the simple additional $1 \times 1$ conv. layers to the pretrained outputs as a linear weighting strategy leads to better results (A and B, E-32 and F). These indicate that a linear combination with $1 \times 1$ conv. layers can really do help for pixelwise prediction.

Third, the detection accuracy improves with the network depth increasing, and saturates when the depth reaches 6 layers (model F). Compared to model A and H, model F can get a higher detection accuracy and the better tradeoff between precision and recall for most cases. It outperforms MATE and ChESS as they produce much false positives. So that the model F is selected as the ultimate detection network and named as RCDN for the further experiments.

*4.2 Simulation on synthetic images*

In order to make an overall evaluation of the detection accuracy of the proposed method, as well as the robustness against to common degenerated factors like noise, image rotation and skew, here we create a set of synthetic images suffered from these factors and investigate their impacts on our corner detector. For benchmarking our model against state-of-the-art methods MATE and ChESS, we apply them to same images respectively without any subpixel refinement, and compare the biases of detected locations to real corners as the criteria.

**Synthetic Images Generation.** In order to make the synthetic images more similar to the real data, we first follow [17] to generate one single corner image by combining four alternating equal-size black and white squares with the intensity of 64 and 191 respectively. Then we insert a transition row and column of 128 at the square borders for simulating the common degenerate scene due to the optical blur and pixel quantization, and getting an integer coordinate of the corners for computing efficiency. An isotropic $3 \times 3$ Gaussian kernel, roughly corresponding to a 0.675 variance, is adopted to the image to simulate the defocus blur. The Gaussian noise with zero mean and standard deviation from 0 to 100 with one step is also added to each pixel. The rotation angle is from 0 to 90 and the skew angle is from 0 to 70. Both of the rotation and skew angle is increased with a step of 1.

**Evaluation Criteria.** Detection accuracy and robustness to these factors can be quantified by measuring the Euclidean distance of each detected location and corresponding ground-truth. The results compared with ChESS and MATE are shown in Fig. 4 and Fig. 5. The coloring of the plots represents the average distance in pixels taken over 1000 trials.

**Evaluation Result.** We can see that RCDN performs better than MATE and ChESS in all experiments. When considered the interferences of noise and rotation, as shown in Fig. 3, RCDN displays higher accuracy and robustness than other two methods. Due to the insufficient learning ability of the shallow layers in MATE. Its performance is slightly worse than the proposed methods. The conventional detector ChESS performs worst, and also presents some periodicity due to the angular spacing of the sampling ring.

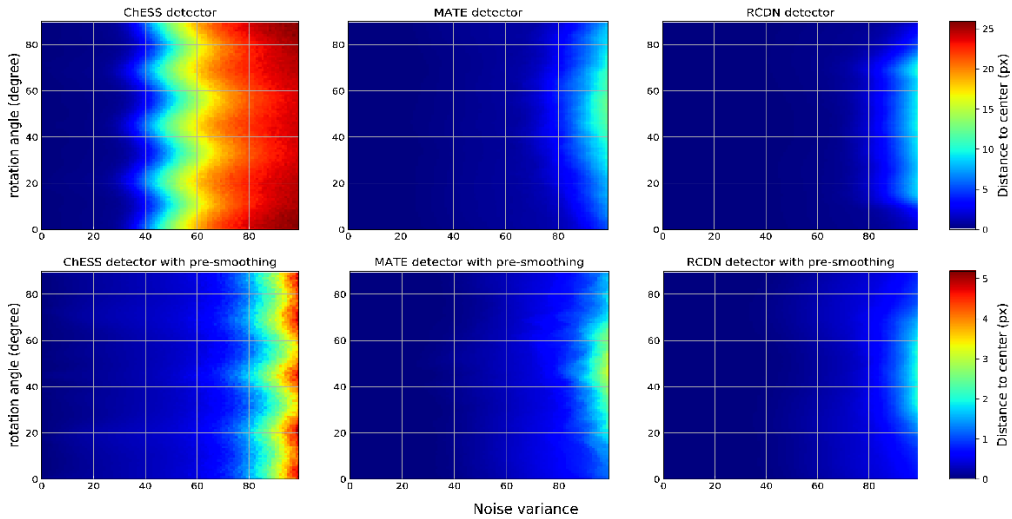

Fig. 4. Comparison of detector performance at various rotation angles and noise levels

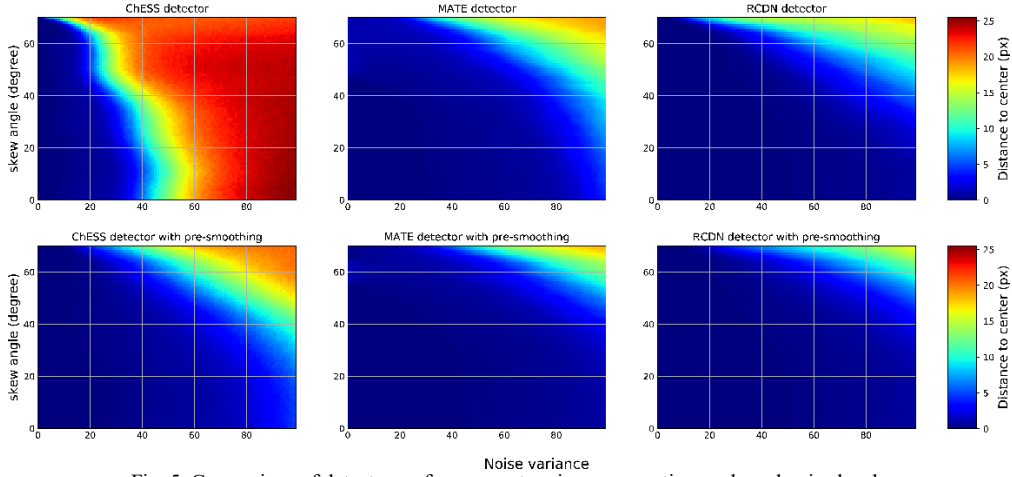
Fig. 5. Comparison of detector performance at various perspective angle and noise levels

The perspective distortion caused by out-of-plane rotation is very common in real-world, which apparently influences the corner appearance by changing the angle of the intersecting lines. Fig. 5 shows the location biases with the skew angle increasing from 0 to 70. RCDN still outperforms other two methods in the same conditions, especially for the higher distortion at all noise levels. MATE have the similar response distribution but is more sensitive to the distortion. ChESS fares worst and cannot get the satisfactory results as the skew angle increases, which give some reason for its poor performance on the GoPro datasets.

Furthermore, it can be seen that the performance of RCDN in these experiments is almost the same, with whether the pre-blur process or not, which is much different from MATE and ChESS. These results show that our RCDN is an excellent X-corner detector for most cases.

### 4.3 Refinement accuracy

For the comprehensive evaluation of our proposed sub-pixel refinement method, here we test response-based methods of CoM (center of mass), Gaussian approximation, parabolic estimator [41], as well as intensity-based methods of edge approximation [31] and surface fitting [6] and compare their performances all on the synthetic corner images undergoing Gaussian Noise, Gaussian blur, image rotation and skew angle. The synthetic images are generated similarly in Sec. 4.2 but with no transition row and column, and all image is randomly subpixel shifted by a fraction of one pixel for precise comparations. The original image is of $80 \times 80$ pixels, but after image renderings, only the center region of $10 \times 10$ pixels is captured for refinement. We would omit the corner candidates not in this region.

The window size is set to $9 \times 9$ pixels for CoM and edge approximation, $5 \times 5$ for surface fitting and $3 \times 3$ for Gaussian approximation and parabolic estimator. The evaluation criteria keep same and the localization errors are averaged from all 1000 trials. When testing one factor's effect, the others are kept at low levels for simulating the real cases.

**Gaussian Noise Test.** As shown in Fig. 5(a), the intensity-based methods diverge form the correct location sharply with the noise increasing, especially for edge approximation. The response-based methods may lead to higher errors at the first but increase slowly, which comply with the result in Sec. 4.2. that the raw corner detector has the higher stability. The proposed method performs best for keeping the lowest errors while maintaining the slowest increase.

**Gaussian Blur Test.** All testing methods can keep stable with the blur altered in Fig. 5(b). This can be well understood that blur conducts just like the common smooth process in post-processing, but not change the profile distribution apparently. Proposed method excels response-based methods largely, while closely approach the intensity-based approach with the acceptable 0.01 pixels bias.

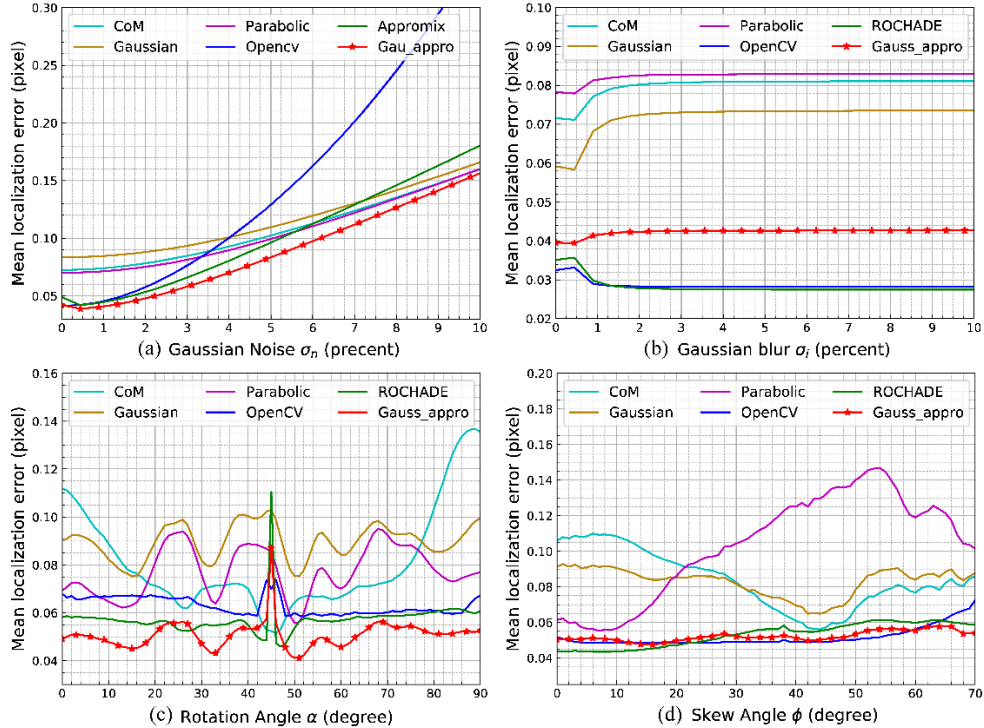

Fig. 5. Mean localization error for different refinement methods with respect to Gaussian noise, Gaussian blur, rotation and skew. OpenCV denotes edge approximation used in OpenCV [33], and ROCHADE denotes surface fitting proposed in ROCHADE [6].

**Rotation Angle Test.** Ideally the location errors keep unchanged with the rotation angle altered. However, due to the discrete sampling the error displays some mild periodicity. In Fig. 5(c), we can see that the phenomenon is distinct for response-based methods, surface fitting method perform better than edge approximation but may get an extremely large error at the angle 45° result from the polynomial fitting limitation. Proposed method may get a less even result but can ease this mutation and get the best result amongst all the rotation.

**Skew Angle Test.** In some low precision photography the out-of-plane rotation is very common to introduce the perspective distortion. Fig. 5(d) illustrates the proposed is similar to the intensity-based methods, they are all more resilient to the image skew than these response-based methods.

All in all, the intensity-based methods perform better than the response-based methods. But the latter could keep an improving stability for image noise due to the robustness of raw corner detector. The proposed integrates the advantages of the two method pipelines, have an desirable advantage especially for the image noise and rotation, and keep almost the same performance for perspective distortion, what's more, with less computational expenses.

### 4.4 Experiments on camera calibration and pose estimation

In this section, we compare the performance of our corner detection method with other state-of-the-art methods in two real-world camera calibration scenarios. The first experiment is taken by the binocular camera setup with 720×1280 resolution mounted on the head-mounted eye tracker and captured 100 views covering a large space. The second was select randomly from the fully visible dataset taken by a wide-angle camera [38]. Example images are illustrated in Fig. 6. Two datasets were evenly split into two series, used for calibration and pose estimation separately. The calibration of MATLAB and OCamCalib were taken with public code, while the others adopted Zhang's method implemented in OpenCV for getting the results.

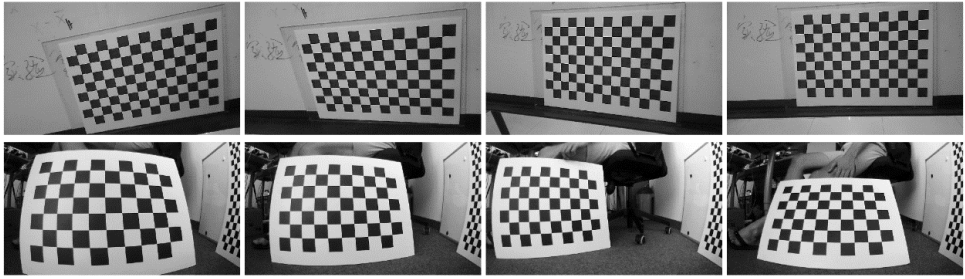

Fig. 6. Sample images of the dataset we captured (the top row) and the dataset from [38] (the second row).

Table 4. Root Mean Squared Error (RMSE) of the reprojection for camera calibration and pose estimation

| Method | Detect Rate | Calibration RMSE [px] | Pose Est. RMSE [px] | Detect Rate | Calibration RMSE [px] | Pose Est. RMSE [px] |
| --- | --- | --- | --- | --- | --- | --- |
| Proposed | 100/100 | **0.17721** | **0.19313** | 100/100 | **0.21387** | **0.21960** |
| OpenCV | 100/100 | 0.18952 | 0.20552 | 100/100 | 0.25041 | 0.24708 |
| MATLAB | 100/100 | 0.18723 | 0.19420 | 100/100 | 0.22613 | 0.22911 |
| Geiger et al. | 100/100 | 0.19018 | 0.20758 | 100/100 | 0.25366 | 0.25287 |
| OCamCalib | 100/100 | 0.29834 | 0.33226 | 100/100 | 0.34511 | 0.46133 |
| Proposed | 85/100 | **0.16958** | **0.18301** | 91/100 | **0.22008** | **0.22017** |
| OCPAD | 85/100 | 0.17017 | 0.18559 | 91/100 | 0.22749 | 0.22743 |

The Root Mean Square Error (RMSE) of the reprojection for camera calibration and pose estimation on two datasets are summarized in Table 4. It is noted that OCPAD fails to find all checkerboard images if all the corners shall be detected. In that case, we apply our methods and OCPAD on the images only successfully detected for a fairer comparison. We can see the proposed methods can get a more satisfactory results over state-of-the-art methods for achieving considerable higher detection rates and a smaller reprojection error.

### 4.5 Extending application of proposed detector

Furthermore, thanks to the intuitive expression of RCDN, it's easily tailored to other types of corners detection, with less fine tuning compared with traditional methods which need human extracted features. Fig. 7 shows three typical examples combined with high-intersection corners. Especially for the corners shown in Fig. 7(b), also called deltille corners, has been discovered desirable for multi-view calibration [1]. We built a small dataset contain 500 deltille images, and used RCDN to train a coarse model, which can get the RMSE of 0.3431 pixel for calibration and 0.3774 pixel for pose estimation on the validation dataset. It shows an exciting potential. More attention will be focus on this modification in the later research.

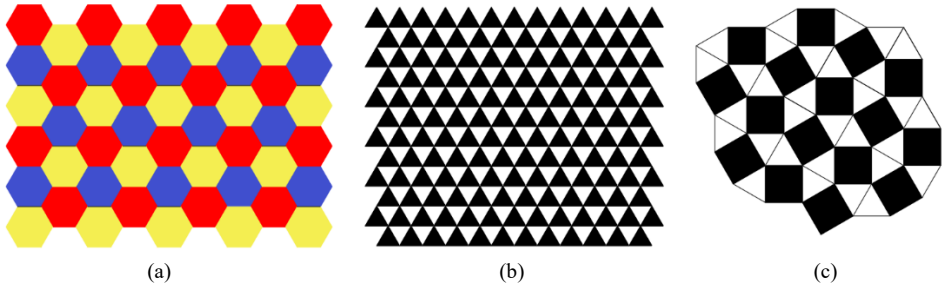

(a)          (b)          (c)

Fig.8. Example patterns with high-order intersection features.

## 5. Conclusion

In this paper we presented a CNN-based detection algorithm (CCDN) to find X-corners, and demonstrated its superior accuracy and robustness to most scenarios. This algorithm mainly contains one X-Corner detection network, and three post-process techniques to select the real corners, it could make a well balance of precision and recall, while maximizing the localization accuracy to various degenerated factors. These advantages will be further enhanced by combing a sub-pixel refinement method, which mixed detection results on pixel intensity and response map, and an improved region growth strategy. This algorithm can also recover the checkerboard structure even not fully visible or partially occluded with high precision X-corner locations. Quantitative comparisons in calibration and pose estimation show it outperforms the state-of-the-art methods. As an efficient X-corner detector, this algorithm can pave the way toward the widespread application of checkerboard pattern in more complexed fields such as outdoors photography, 3D reconstruction, and motion-tracking.

## Declaration of Competing Interest

The authors declare no conflicts of interest.

## Acknowledgements

This work is partially supported by the National Natural Science Foundation of China (Grant No. U1913601 and No. 91648203) and the International Science & Technology Cooperation Program of China (Grant No. 2016YFE0113600).